\documentclass[10pt,twocolumn,letterpaper]{article}

\usepackage{cvpr}
\usepackage{times}
\usepackage{epsfig}
\usepackage{graphicx}
\usepackage{amsmath}
\usepackage{amssymb}

\usepackage[breaklinks=true,bookmarks=false]{hyperref}

\cvprfinalcopy % *** Uncomment this line for the final submission

\def\cvprPaperID{****} % *** Enter the CVPR Paper ID here

\begin{document}

%%%%%%%%% TITLE
\title{Webpage Saliency Prediction with Two-stage Generative Adversarial Networks }

\author{Yu Li\\
Shanghai Jiao Tong University\\
Shanghai, China\\
{\tt\small liyu\_sjtu@sjtu.edu.cn}
\and
Ya Zhang\\
Shanghai Jiao Tong University\\
Shanghai, China\\
{\tt\small ya\_zhang@sjtu.edu.cn}
}

\maketitle
%\thispagestyle{empty}

%%%%%%%%% ABSTRACT
\begin{abstract}
   Web page saliency prediction is a challenge problem in image transformation and computer vision. In this paper, we propose a new model combined with web page outline information to prediction people's interest region in web page. For each web page image, our model can generate the saliency map which indicates the region of interest for people. A two-stage generative adversarial networks are proposed and image outline information is introduced for better transferring. Experiment results on FIWI dataset show that our model have better performance in terms of saliency prediction.
\end{abstract}

%%%%%%%%% BODY TEXT
\section{Introduction}

With the wide spread of Internet and the development of e-commerce, online social networking, and search engineering in recent years, web page has played an important role in our daily life. According to the statics data on Internet Live Stats Online, there are more than 1.8 billion websites exsiting on the worlds by the end of May,2018. According to the statics data on Telecommunication Union (ITU), the number of Internet user on the earth has been more than 3.3 billion by the end of 2016. Web pages affect people's lives all the time. 

The market of Internet is so huge and how to design a attractive web has been the first priority. A good web page, not only need to meet user's normal functional requirements, but also should be able to grasp the user's eyeballs using reasonable layout and typography. Web designer can display high-yieding products in the area with more user's attention and display low-yielding products in the area with less user's attention.
\begin{figure}[!th]
  \centering
  \includegraphics[width=7cm]{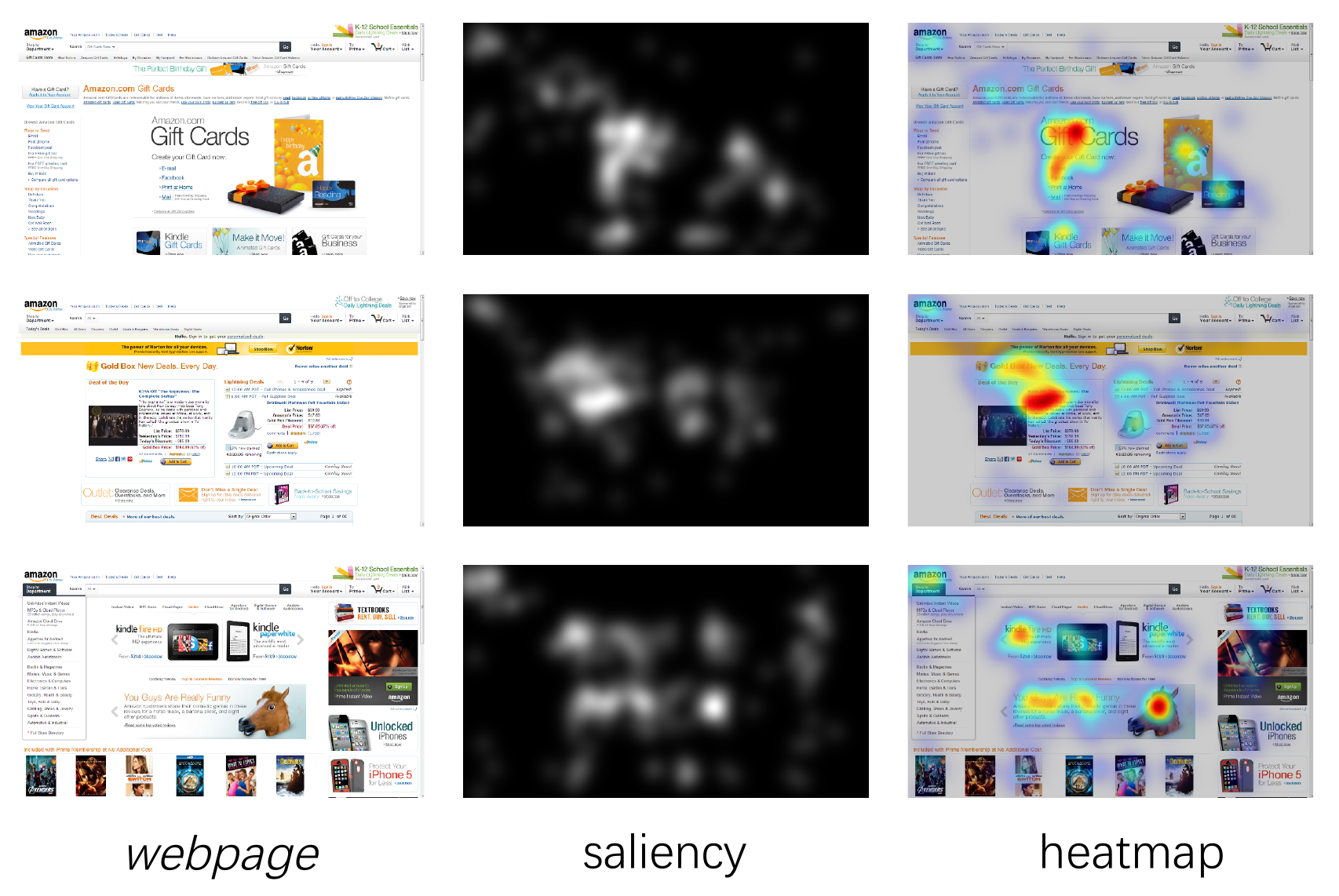}\\
  \caption{Human eye tracking on web page images}\label{intro}
  \vspace{-13pt}
\end{figure}

So how to effectively predict the saliency maps of web images(Fig.\ref{intro}) become a very important problem. In recent years, there has been a lot of good works in predicting the saliency of natural images. However, unlike natural images, web page images have more texts, logos and animations, which are rich in outline information and not exist in natural images. Besides, the natural image is more relevant in each area. For example, there are often green lands next to the trees, and there are often blue sky over the lake. But for the web page images, each area block is often independent, which increases the difficulty of web page saliency prediction for one time.

Taking into account the above problems, we design a two-stage generative adversarial networks(TSGAN), allowing the model to achieve a two-step prediction of web page saliency.
Firstly, in the first step of prediction, our model takes the original web page images as input and uses a encoder-decoder to generate a corase prediction saliency maps. Secondly, in the second step of prediction, the corase prediction results obtained in the first step and the web pages are stitched and sent to the network, and the encoder-decoder is used again to get a fine prediction. In addition, in the two-step encoder-decoder structure, we introduce the outline information of the image extracted by the Laplacian operator to further improve the network's learning ability. In order to better demonstrate the capabilities of our model, we designed a series of experiments based on FIWI\cite{shen2015predicting}. The experimental results show that our model is not only perform well in visual effects, but also excellent in objective metrics.

The main contributions of our work are:

1. We present TSGAN to achieve two-step prediction for web page images and achieve very good results.

2. We introduce the outline information in the process of web page prediction for the first time and improve the accuracy of prediction.

3. Experimental results show that our method outperforms state of the art model results
%-------------------------------------------------------------------------
\section{Related Works}
\subsection{Saliency Prediction on Natural Images}
Over several years, works on saliency detection mainly focus on natural images because the conspicuity of region in natural images are easier to distinguish than other images. Some early models used low-level features such as intensity, edge orientations and color contrast to predict saliency.  Itti \cite{Itti2000A} proposed a method that combined color, orientation and intensity information by center-surround difference and normalization. Schlopf presented GBVS model~\cite{Sch2006Graph} based on graph model and combined with Markov chain. With the development of machine learning, there are some methods that adopted sparse coding and independent component analysis(ICA) to learn low-level features. ICL~\cite{Hou2009Dynamic}, SUN~\cite{Zhang2008SUN} and  AIM~\cite{Bruce2009Saliency} extracted features by ICA. Itti and Borji~\cite{Borji2012Exploiting} presented their model using sparse coding to learn feature.

\section{Two-stage generative adversarial networks}
Although low-level feature were effectively used in saliency prediction, high-level feature were explored to predict fixations. Recently advances of convolutional neural network(CNN) provides powerful assistance for fixations prediction. CNN can effectively extract the high-level features from large training data. Huang~\cite{Huang2015SALICON} proposed a deep neural network(DNN) architecture for saliency prediction in which CNN was used as feature extractor. Pan proposed Shallow Convnet and Deep Convnet~\cite{Pan2016Shallow} model based on pre-trained VGG network.

\subsection{Saliency Prediction on Web page Images}
Webpage, shown on computer browser, are being contacted by more and more people on internet. Different from natural images, web page images have plenty of text information which is full of semantic information.  In recent years, attention models began to use for web page saliency prediction. Faraday proposed visual scanning model~\cite{Faraday2000Visually} which use visual attention in web page for the first time. Girer $et$ $al.$ presented three perceptual attentional mechanisms~\cite{Grier2007How} including to "top-left corner of whole web page is most noticeable", "overly striking area in web page do not contain information" and "information with same type will be got together". These theory explains the viewer's behavior on webpage.

Although attention model played important role in web page saliency prediction, there have been lots of work based on layout of web page. Bendersky\cite{Bendersky2011Quality} proposed a method that ranked web page based on readability, layout and so on. In their model, taking the web page layout into consideration, the model had better performance than other models. Pirlo et al.\cite{Pirlo2013Layout} presented a layout-based document-retrieval System and extracted layout-based features. The effectiveness of model also showed that layout information can not be ignored in web page saliency prediction

\subsection{Generative Adversarial Networks}
Goodfellow proposed the novel framework called Generative Adversarial Networks (GANs)~\cite{Goodfellow2014Generative} which consists of a generator ($G$) and a discriminator ($D$) and is to learn generative models. The generator maps a noisy distribution into a data distribution and generates data as real as possible to fool the discriminator. At the same time, the discriminator learns to distinguish the generated data from the real data and not to be fooled by the generator.  Both the generator and discriminator are deep neural networks. 
Taking a noise vector $z$ as input which is sampled from the uniform distribution [-1,1], the generator synthesizes a fake image $G(z)$. Then the discriminator calculates the probability that the synthesized image is from real images. The whole objective of the GAN can be written as:
\begin{equation}
\min_G\mathbb{E}_{z\sim p_z}[log(1-D(G(z)))]
\end{equation}
\begin{equation}
\max_D\mathbb{E}_{x\sim p_x}[logD(x)]+\mathbb{E}_{z\sim p_z}[log(1-D(G(z)))]
\end{equation}

where $p_z$ and $p_x$ represent the distribution of noise vector and real data respectively. The GAN framework corresponds to a zero-sum game, if both discriminator
and generator work well enough, $D(x_{fake})$ = $D(x_{real})$ = 0.5.

So far, lots of variations of GANs have been proposed and applied in image generation~\cite{pmlr-v70-arjovsky17a}, image inpainting~\cite{Yijun2017Generative}, super resolution~\cite{Ledig_2017_CVPR}, style transfer~\cite{DBLP:journals/corr/ZhangD17}.
Arjovsky proposed the Wasserstein GAN (WGAN)~\cite{pmlr-v70-arjovsky17a} to improve the stability of GAN, in which the KL divergence was replaced by the Earth-Mover (EM) distance. They also provided a rigorous theoretical analysis about how EM distance works. Due to the good stability of WGAN, we construct the adversarial loss based on WGAN loss in our study. Ledig presented super-resolution generative adversarial networks (SRGAN)~\cite{Ledig_2017_CVPR} to reconstruct realistic texture details, where the residual blocks~\cite{He2016Deep} were employed both in generator and discriminator for better results. Zhang combined GANs and style transfer ~\cite{Gatys2016Image} based on pre-trained VGG network. Instead of iterating the synthesized images on pixel level, they updated the parameters directly and got an explicit result.
%------------------------------------------------------------------------
\begin{figure*}[!th]
  \centering
  \includegraphics[width=18cm]{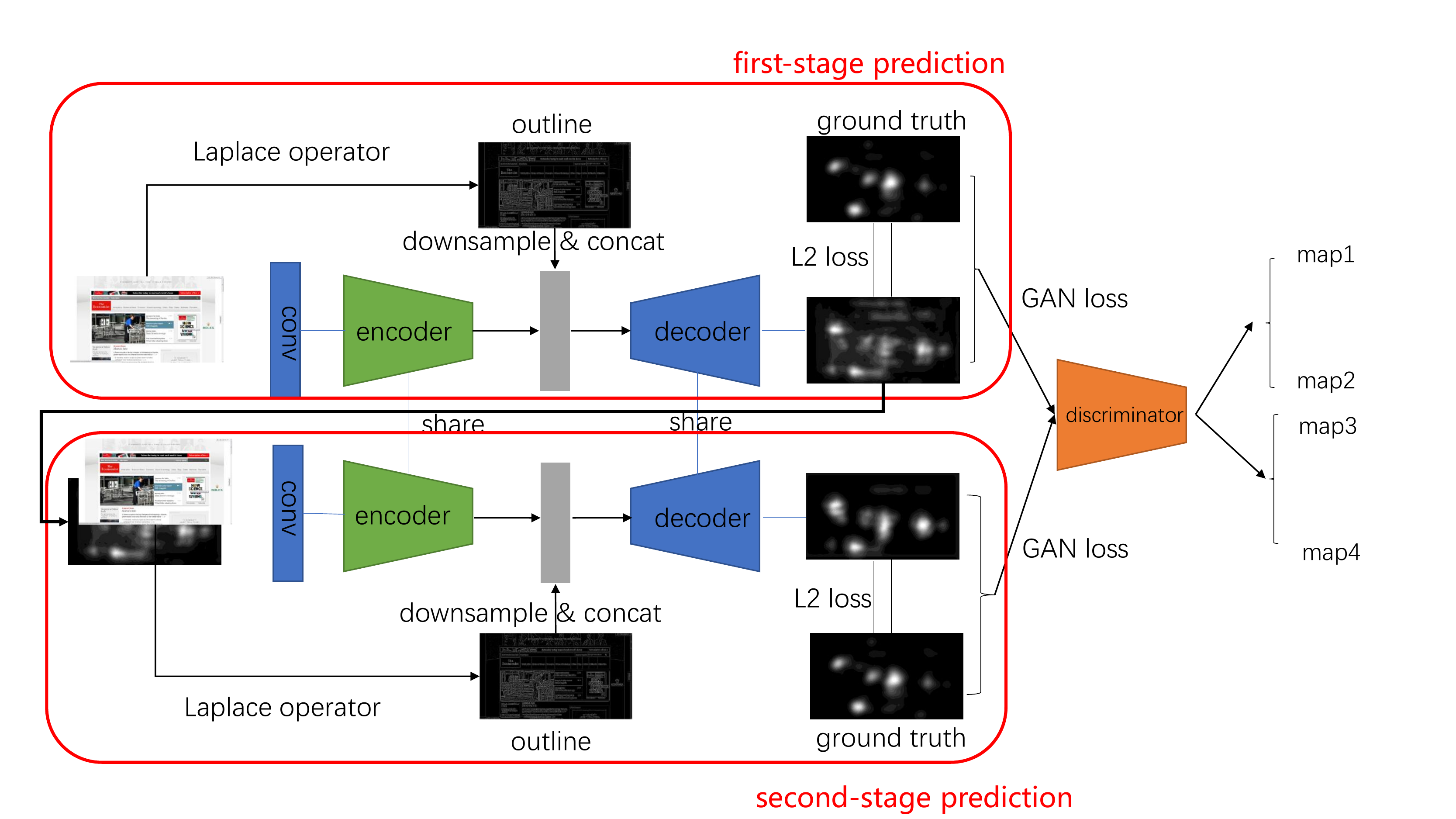}\\
  \caption{The overall architecture of TSGAN. }\label{model}
  \vspace{-13pt}
\end{figure*}
\section{Two-stage generative adversarial networks}

In this section, we present our two-stage generative adversarial networks(TSGAN), whose architecture is shown in Fig.~\ref{model}. 

The main novelty of our model is dividing saliency prediction process into two steps. At first prediction stage, we adopt encoder-decoder model combined with generative adversarial nets to predict the saliency region and take web page image as input data. At the second prediction stage, we reuse the model parameters both in encoder and decoder except for the first convolutional layer, because at the second stage, we take both web page image and croase saliency map which predicted by first prediction stage. 

\subsection{Encoder}
Encoder take image as input to extract features which can represent high level feature which can not be shown in image pixel. Our encoder model make use of 8 convolutional layers and 4 max pooling layers with stride of 2, and output feature passed through convolutional layers should be operated by relu function. In order to accelerate model convergence speed, we initialize encoder parameters by VGG networks\cite{}, which is pre-trained in ImageNet dataset for classification task. In summary, the detail of encoder structure is illustrated in Fig.\ref{parts_structure}

\subsection{Decoder}

On the contrary of encoder, decoder aims to transfer high level features into pixel level images. Compared with natural images, web page images have more lines and blocks which can be demonstrated by outline information.
Considering the particularity of web page images, we downsample the outline images which are extracted by laplace of gaussian operator\cite{Marr1980Royal} for edge dectection, and concatenate into feature maps extracted by encoder as input of decoder. Besides, we introduce dilated residual block, which contains two dilated convolutional layers, two batch normalization and input map will be added by output map. In decoder model, we make use of 8 dilated residual blocks and 4 deconvolutional layers.
The detail if decoder structure is shown in Fig. \ref{parts_structure}
\subsection{Discriminator}

\begin{figure*}[!th]
  \centering
  \includegraphics[width=18cm]{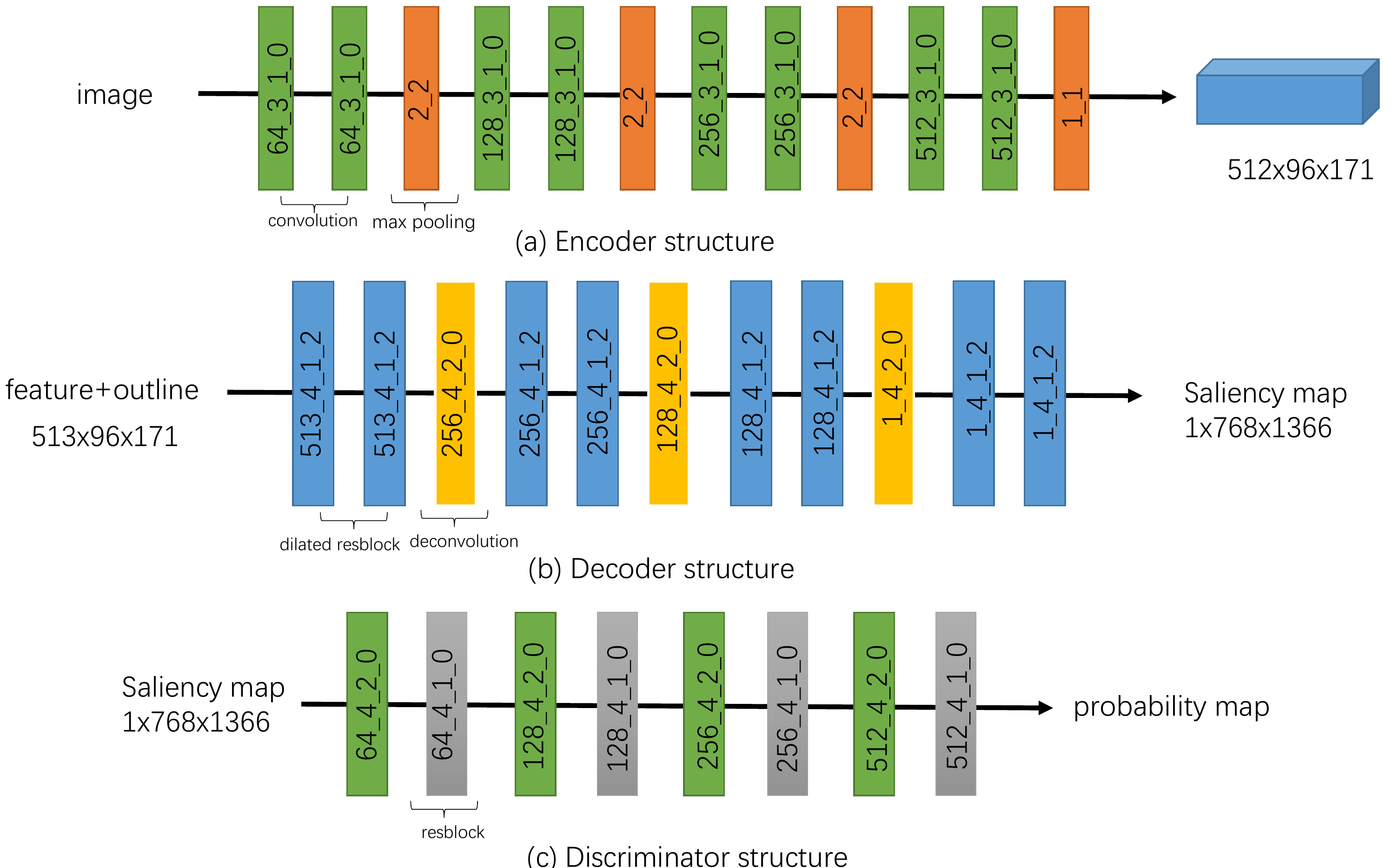}\\
  \caption{The architecture of encoder, decoder and discriminator. Every convolutional layer and block are expressed as $channle\_kernel\_stride\_holes$ and pooling layer is expressed as $kernel\_stride$}\label{parts_structure}
  \vspace{-13pt}
\end{figure*}

As we have explained in related work section, we utlize the idea of GAN to constraint web page saliency prediction. We treat the encoder-decoder part as generator of GANs, and discriminator is an important part which distinguishes the real saliency images and predicted saliency images. Different from traditional discriminator in GANs, our discriminator predicts a probability map.
$D$ is composed of 8 convolutional layers without any fully connected layer.At the two stages prediction process, the discriminator is totally shared.  The detail of discriminator is shown in Fig.\ref{parts_structure}.

\subsection{Loss Function}
The goal of our model is to realize web page image saliency prediction, so some loss constraints the saliency map generation.

\textbf{At the first prediction stage:} The first encoder-decoder($G_1$) predict a coarse saliency map with the source web images as input. L2 loss is used as supervised loss, which is defined as :
\begin{equation} \label{eq:1}
\L_{1}(x, s) = \sum_{c=1}^{C_l}\sum_{h=1}^{H_l}\sum_{w=1}^{W_l}\frac{(s_{i,j}-G_1(x)_{i,j})^2}{2W_lH_lC_l}
\end{equation}
Where x denotes the web page image, s denotes ground truth saliency map and $C_l$, $W_l$, $H_l$ respectively represent dimensions of saliency map.

GAN loss is usually used in image generation and image translation, and has been achieved good performance. The generator and discriminator try to do a oppesite task, generator tries to synthesize fake image to fool the discriminator, and the discriminator tries to distinguish the fake image and the real image. GAN loss can be written as:
\begin{equation}\label{eq:g}
\begin{split}
\L_{2}(G_1,D)=&\min_{G_1} \max_{D}\mathbb{E}_{s\sim P(s)}[D(s)] + \\
&\mathbb{E}_{x\sim P(x)}[1-D(G_1(x))]
\end{split}
\end{equation}

\textbf{At the second prediction stage:} The second encoder-decoder($G_2$) predict fine saliency map with source web page image and coarse saliency map predicted by first stage nets as input. We also adote L2 loss and GAN loss for training. The equation is shown below:
\begin{equation} \label{eq:3}
\L_{3}(x,\hat{s}, s) = \sum_{c=1}^{C_l}\sum_{h=1}^{H_l}\sum_{w=1}^{W_l}\frac{(s_{i,j}-G_2(x,\hat{s})_{i,j})^2}{2W_lH_lC_l}
\end{equation}

\begin{equation}\label{eq:4}
\begin{split}
\L_{4}(G_2,D)=&\min_{G_2} \max_{D}\mathbb{E}_{s\sim P(s)}[D(s)] + \\
&\mathbb{E}_{x\sim P(x)}[1-D(G_2(x,G_2(x))]
\end{split}
\end{equation}

Besides L2 loss and GAN loss, we also employ the total variation regularizer\cite{DBLP:journals/corr/ZhangD17} to smooth the final results.

\begin{equation}
\L_{tv}^\alpha(x)=\sum_{i,j}\left((x_{i,(j+1)}-x_{i,j})^2+(x_{(i+1),j}-x_{i,j})^2\right)^{\frac{1}{\alpha}}
\end{equation}

In summary, the final objective function can be as:
\begin{equation}
\begin{split}
\L(G_1,G_2,D)=&\lambda_1L_{1}+\lambda_2L_{2}+\lambda_3L_{3}+\lambda_4L_{4}+\lambda_5L_{tv}^{\alpha}
\end{split}
\end{equation}
where ($\lambda_1$,$\lambda_2$,$\lambda_3$,$\lambda_4$,$\lambda_5$) are the trade-off weights for different loss parts.
\begin{figure*}[!th]
  \centering
  \includegraphics[width=18cm]{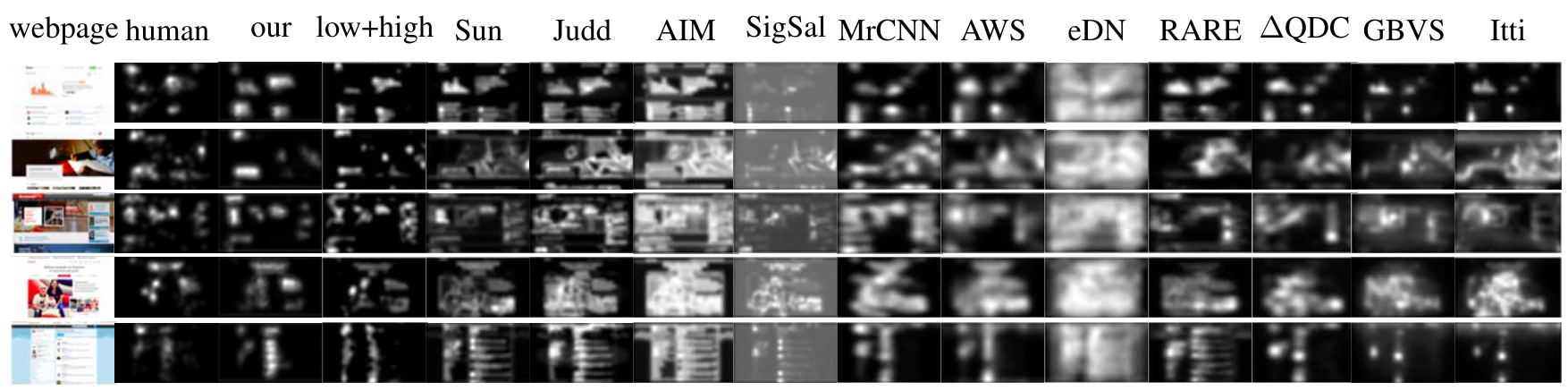}\\
  \caption{TThe results for different models on FiWI data.}\label{comprason}
  \vspace{-13pt}
\end{figure*}
\section{Experiments and Results}
In this section, we show the performance of our model in FIWI dataset\cite{shen2014webpage} and compare our two-stage generative adversarial nets with several baseline models. Besides, in order to show the effect of outline information and two-stage prediction, we do self-contrast experiments and analyze the results in Section.

In our experiments, we set the trade-off parameters $\lambda_1=0.1$, $\lambda_2=1$, $\lambda_1=0.1$, $\lambda_2=1$, $\lambda_1=0.1$.

\subsection{Dateset}
The FiWI dataset contains 149 web page screenshots with eye movement data from 11 observers during free-viewing. Different from natural images, web page screenshots contain lots of edge information such as much words and blocks.
FIWI is divided into 50 textual images, 50 pictorial images and 49 mixed images according to the proportion of words and pictures.
We divide FIWI data into training data which contains 123 screenshots and test data which contains 26 screenshots.

%-------------------------------------------------------------------------
\subsection{Performance Comparison}
In this section, we first compare our model with several baseline models, including Itti\cite{Itti2000A}, GBVS\cite{harel2007graph}, $\Delta$QDCT\cite{schauerte2012predicting}, RARE\cite{riche2013rare2012}, eDN\cite{vig2014large}, AWS\cite{garcia2012relationship}, MrCNN\cite{liu2015predicting}, SigSal\cite{hou2012image}, AIM\cite{Bruce2009Saliency}, Judd\cite{judd2009learning}, SUN\cite{Zhang2008SUN}, Low+High\cite{shen2015predicting}. All these models are conducted in FIWI dataset and our model has a better performance as Fig.\ref{comprason}.

The saliency map evaluation metrics we use are linear Correlation Coefficient(CC) and Normalized Scanpath Saliency(NSS). CC meatures the linear correlations between synthesized saliency map and ground truth saliency map. The predicted saliency can be regarded same as ground truth if CC equal to 1. NSS score indicates the average value of the normalized saliency map in the fixation points. The larger NSS score is, the better the model works.
The performance of all model can be seen in Table\ref{metrics}

\begin{figure}[!th]
  \centering
  \includegraphics[width=8.5cm]{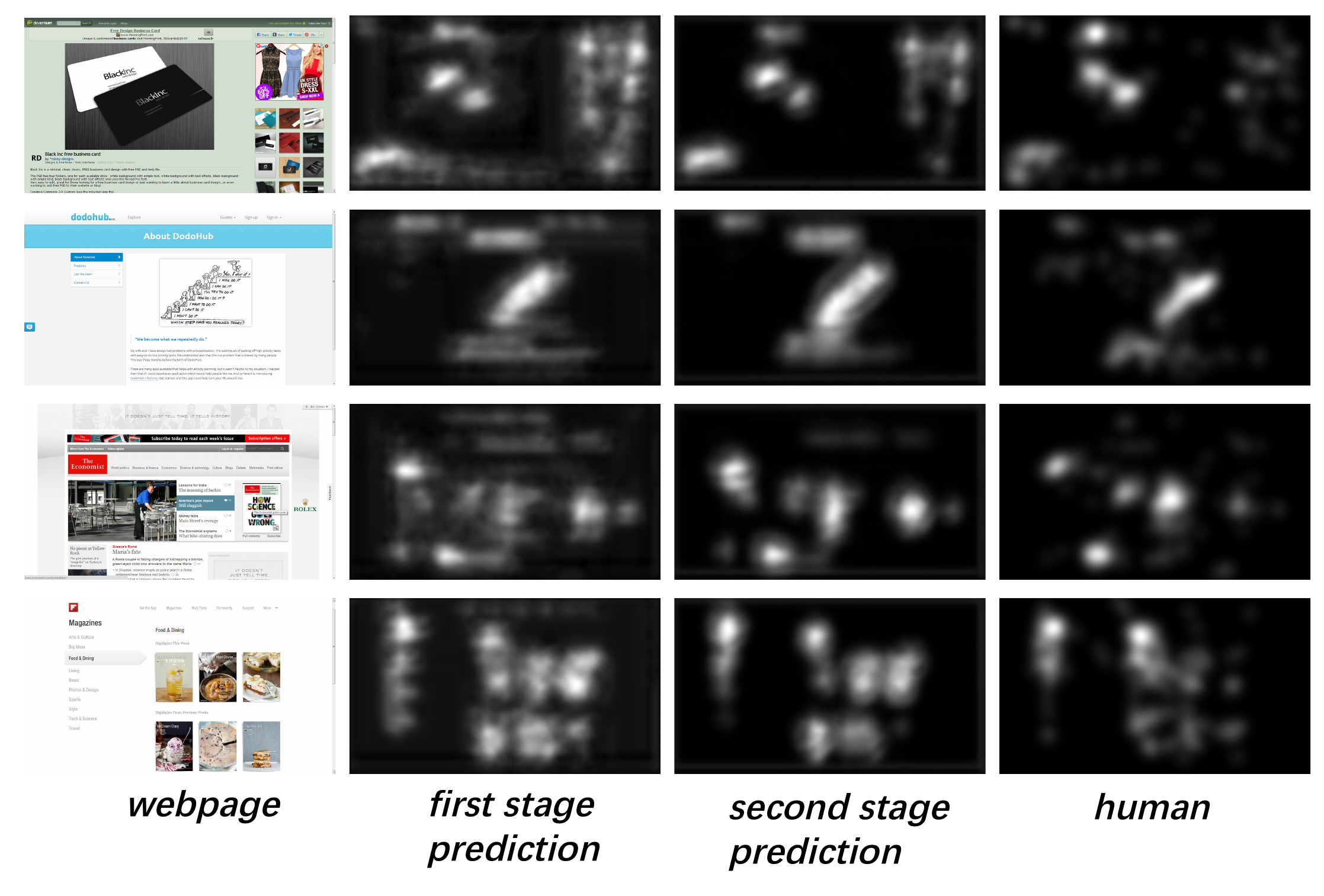}\\
  \caption{The comparison between the two predicted saliency maps by our model. From the left to right, the images are web page images, saliency map predicted by first stage, saliency maps predicted by second stage and human saliency maps, respective. }\label{self-constrast}
  \vspace{-13pt}
\end{figure}

%-------------------------------------------------------------------------
\subsection{Algorithmic analysis}

In this section, We discuss the validity of our model.

\begin{figure}[!th]
  \centering
  \includegraphics[width=8.5cm]{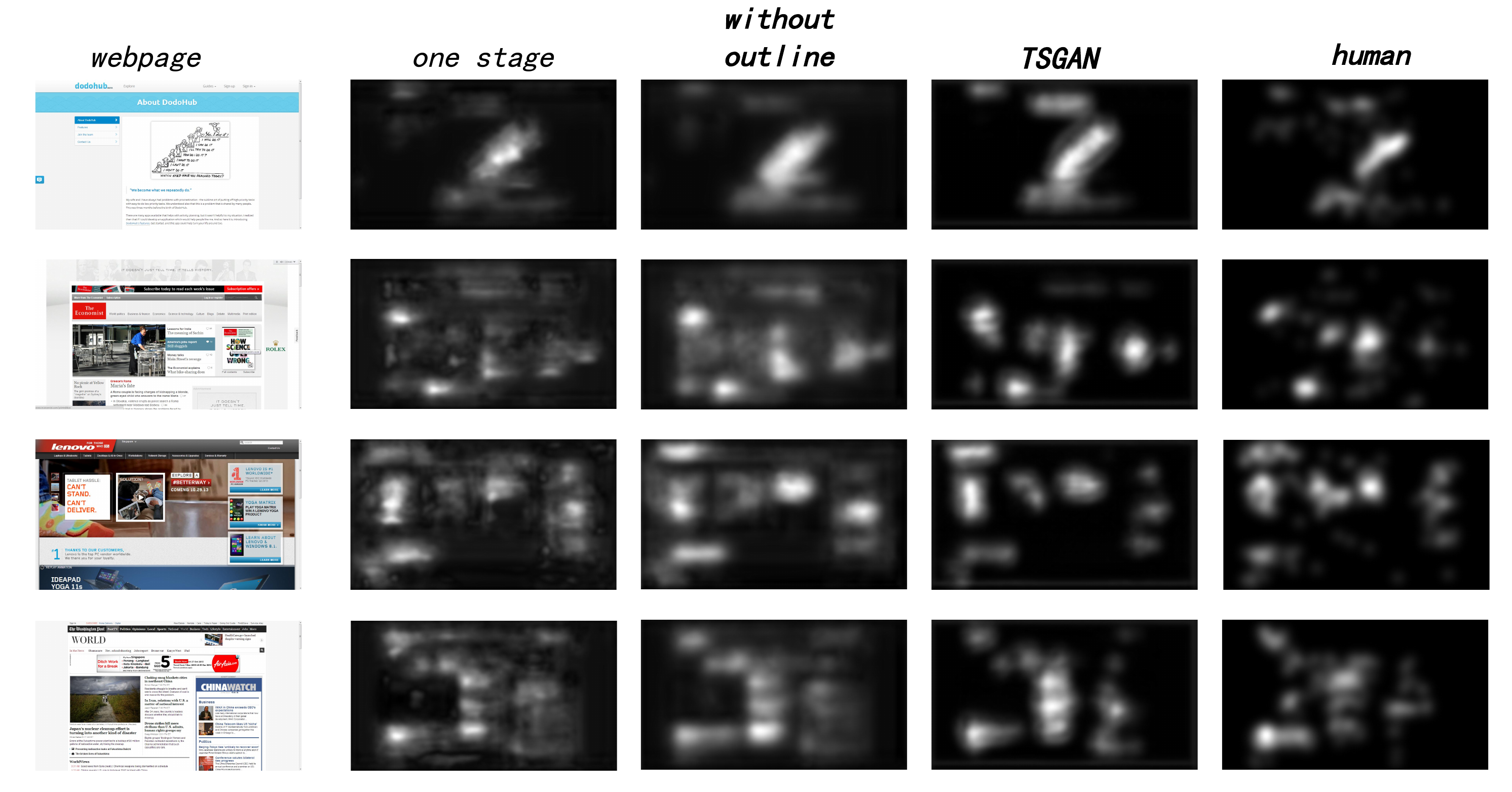}\\
  \caption{The comparison between each parts of TSGAN. From the left to right, the images are web page images, saliency map predicted by model which has only one stage prediction, saliency maps predicted by TSGAN without outline information and human saliency maps, respective. }\label{outline}
  \vspace{-13pt}
\end{figure}

First, we analyze the outline information and two-stage structure to see whether they play an important role in model training. We conduct three experiments: removing the outline information but keeps other structure unchanged; removing the second prediction stage and taking first stage prediction as final saliency map. The results of self-constraint experiments can be seen in Fig.\ref{outline}

Second, we demonstrate the improvement between the two-stage prediction. We compare the first-stage prediction results and the second-stage prediction in Fig.\ref{self-constrast}. And we can find that the first-stage prediction is obviously blurred in the background.
\begin{table}
\begin{center}
\caption{Classification accuracy on mouth close/open and mustache add/removal tasks.}\label{metrics}
\begin{tabular}{|l|c|c|}
\hline
Method & CC & NSS \\\hline
%\hline\hline
Itti & 0.2312 & 0.4626 \\
GBVS & 0.3385 &  0.6893 \\
$\Delta$QDCT & 0.3018 & 0.6095 \\
RARE & 0.3598 & 0.7580 \\
eDN & 0.4238 &  0.8507 \\
AWS & 0.4002 &  0.8421 \\
MrCNN & 0.4313 & 0.9207 \\
SigSal & 0.4112 & 0.8840 \\
AIM & 0.4460 &  0.9290 \\
Judd & 0.4676 &  1.0168 \\
SUN & 0.4789 & 1.0274 \\
Low+High & 0.5947 & 1.3578 \\
Ours & 0.6926 & 1.5028 \\
\hline
\end{tabular}
\end{center}
% \vspace{-13pt}
\end{table}

%------------------------------------------------------------------------
\section{Conclusion}

The results in our paper suggest that two-stage GAN model performs well in web page saliency prediction task and both outline information and two-stage structure play important role in the entire model. web page outline information can be utilized sufficiently to saliency prediction and has been achieved significant improvement. To train the two-stage generative networks, we design three parts loss for both prediction stages to force the generator gradient to update in right direction. Two L2 loss keep the prediction saliency map close to ground truth saliency map, two adversarial loss aim make final saliency more real, total variation regularization smooth the final images.

Although our model has achieved outstanding performance in web page saliency prediction, we do not transfer our model into natural images. Limited to the number of dataset, our results are not totally accurate compared with ground truth.

In the future work, we will do some experiments on model transfer into natural images and few-shot learning.

{\small
\bibliographystyle{ieee}
\bibliography{egbib}

\begin{thebibliography}{10}\itemsep=-1pt

\bibitem{pmlr-v70-arjovsky17a}
M.~Arjovsky, S.~Chintala, and L.~Bottou.
\newblock Wasserstein generative adversarial networks.
\newblock In {\em Proceedings of the 34th International Conference on Machine
  Learning}, volume~70, pages 214--223. PMLR, 06--11 Aug 2017.

\bibitem{Bendersky2011Quality}
M.~Bendersky, W.~B. Croft, and Y.~Diao.
\newblock Quality-biased ranking of web documents.
\newblock In {\em Acm International Conference on Web Search \& Data Mining},
  pages 95--104, 2011.

\bibitem{Borji2012Exploiting}
A.~Borji and L.~Itti.
\newblock Exploiting local and global patch rarities for saliency detection.
\newblock In {\em Computer Vision and Pattern Recognition}, pages 478--485,
  2012.

\bibitem{Bruce2009Saliency}
N.~D.~B. Bruce and J.~K. Tsotsos.
\newblock Saliency, attention, and visual search: An information theoretic
  approach.
\newblock {\em Journal of Vision}, 9(3):1--24, 2009.

\bibitem{Faraday2000Visually}
P.~Faraday.
\newblock {\em Visually Critiquing Web Pages}.
\newblock Springer Vienna, 2000.

\bibitem{garcia2012relationship}
A.~Garcia-Diaz, V.~Leboran, X.~R. Fdez-Vidal, and X.~M. Pardo.
\newblock On the relationship between optical variability, visual saliency, and
  eye fixations: A computational approach.
\newblock {\em Journal of vision}, 12(6):17--17, 2012.

\bibitem{Gatys2016Image}
L.~A. Gatys, A.~S. Ecker, and M.~Bethge.
\newblock Image style transfer using convolutional neural networks.
\newblock {\em IEEE Conference on Computer Vision and Pattern Recognition},
  pages 2414--2423, 2016.

\bibitem{Goodfellow2014Generative}
I.~J. Goodfellow, J.~Pouget-Abadie, M.~Mirza, B.~Xu, D.~Warde-Farley, S.~Ozair,
  A.~Courville, and Y.~Bengio.
\newblock Generative adversarial nets.
\newblock pages 2672--2680, 2014.

\bibitem{Grier2007How}
R.~A. Grier, P.~Kortum, and J.~T. Miller.
\newblock How users view web pages: An exploration of cognitive and perceptual
  mechanisms.
\newblock {\em Human Computer Interaction Research in Web Design \&
  Evaluation}, 2007.

\bibitem{harel2007graph}
J.~Harel, C.~Koch, and P.~Perona.
\newblock Graph-based visual saliency.
\newblock In {\em Advances in neural information processing systems}, pages
  545--552, 2007.

\bibitem{He2016Deep}
K.~He, X.~Zhang, S.~Ren, and J.~Sun.
\newblock Deep residual learning for image recognition.
\newblock In {\em Computer Vision and Pattern Recognition}, pages 770--778,
  2016.

\bibitem{hou2012image}
X.~Hou, J.~Harel, and C.~Koch.
\newblock Image signature: Highlighting sparse salient regions.
\newblock {\em IEEE transactions on pattern analysis and machine intelligence},
  34(1):194--201, 2012.

\bibitem{Hou2009Dynamic}
X.~Hou and L.~Zhang.
\newblock Dynamic visual attention: Searching for coding length increments.
\newblock In {\em Conference on Neural Information Processing Systems,
  Vancouver, British Columbia, Canada, December}, pages 681--688, 2009.

\bibitem{Huang2015SALICON}
X.~Huang, C.~Shen, X.~Boix, and Q.~Zhao.
\newblock Salicon: Reducing the semantic gap in saliency prediction by adapting
  deep neural networks.
\newblock In {\em IEEE International Conference on Computer Vision}, pages
  262--270, 2015.

\bibitem{Itti2000A}
L.~Itti and C.~Koch.
\newblock A saliency-based search mechanism for overt and covert shifts of
  visual attention.
\newblock {\em Vision Research}, 40(12):1489--1506, 2000.

\bibitem{judd2009learning}
T.~Judd, K.~Ehinger, F.~Durand, and A.~Torralba.
\newblock Learning to predict where humans look.
\newblock In {\em Computer Vision, 2009 IEEE 12th international conference on},
  pages 2106--2113. IEEE, 2009.

\bibitem{Ledig_2017_CVPR}
C.~Ledig, L.~Theis, F.~Huszar, J.~Caballero, A.~Cunningham, A.~Acosta,
  A.~Aitken, A.~Tejani, J.~Totz, Z.~Wang, and W.~Shi.
\newblock Photo-realistic single image super-resolution using a generative
  adversarial network.
\newblock In {\em The IEEE Conference on Computer Vision and Pattern
  Recognition (CVPR)}, July 2017.

\bibitem{Yijun2017Generative}
Y.~Li, S.~Liu, J.~Yang, and M.-H. Yang.
\newblock Generative face completion.
\newblock {\em IEEE Conference on Computer Vision and Pattern Recognition},
  2017.

\bibitem{liu2015predicting}
N.~Liu, J.~Han, D.~Zhang, S.~Wen, and T.~Liu.
\newblock Predicting eye fixations using convolutional neural networks.
\newblock In {\em Computer Vision and Pattern Recognition (CVPR), 2015 IEEE
  Conference on}, pages 362--370. IEEE, 2015.

\bibitem{Marr1980Royal}
D.~Marr and E.~Hildreth.
\newblock Theory of edge detection.
\newblock {\em Biological Sciences}, 1980.

\bibitem{Pan2016Shallow}
J.~Pan, E.~Sayrol, X.~Giroinieto, K.~Mcguinness, and N.~E. Oconnor.
\newblock Shallow and deep convolutional networks for saliency prediction.
\newblock In {\em Computer Vision and Pattern Recognition}, pages 598--606,
  2016.

\bibitem{Pirlo2013Layout}
G.~Pirlo, M.~Chimienti, M.~Dassisti, D.~Impedovo, and A.~Galiano.
\newblock {\em Layout-Based Document-Retrieval System by Radon Transform Using
  Dynamic Time Warping}.
\newblock Springer Berlin Heidelberg, 2013.

\bibitem{riche2013rare2012}
N.~Riche, M.~Mancas, M.~Duvinage, M.~Mibulumukini, B.~Gosselin, and T.~Dutoit.
\newblock Rare2012: A multi-scale rarity-based saliency detection with its
  comparative statistical analysis.
\newblock {\em Signal Processing: Image Communication}, 28(6):642--658, 2013.

\bibitem{schauerte2012predicting}
B.~Schauerte and R.~Stiefelhagen.
\newblock Predicting human gaze using quaternion dct image signature saliency
  and face detection.
\newblock In {\em Applications of Computer Vision (WACV), 2012 IEEE Workshop
  on}, pages 137--144. IEEE, 2012.

\bibitem{Sch2006Graph}
B.~Schölkopf, J.~Platt, and T.~Hofmann.
\newblock Graph-based visual saliency.
\newblock In {\em International Conference on Neural Information Processing
  Systems}, pages 545--552, 2006.

\bibitem{shen2015predicting}
C.~Shen, X.~Huang, and Q.~Zhao.
\newblock Predicting eye fixations on webpage with an ensemble of early
  features and high-level representations from deep network.
\newblock {\em IEEE Transactions on Multimedia}, 17(11):2084--2093, 2015.

\bibitem{shen2014webpage}
C.~Shen and Q.~Zhao.
\newblock Webpage saliency.
\newblock In {\em European conference on computer vision}, pages 33--46.
  Springer, 2014.

\bibitem{vig2014large}
E.~Vig, M.~Dorr, and D.~Cox.
\newblock Large-scale optimization of hierarchical features for saliency
  prediction in natural images.
\newblock In {\em Proceedings of the IEEE Conference on Computer Vision and
  Pattern Recognition}, pages 2798--2805, 2014.

\bibitem{DBLP:journals/corr/ZhangD17}
H.~Zhang and K.~Dana.
\newblock Multi-style generative network for real-time transfer.
\newblock {\em arXiv preprint arXiv:1703.06953}, 2017.

\bibitem{Zhang2008SUN}
L.~Zhang, M.~H. Tong, T.~K. Marks, H.~Shan, and G.~W. Cottrell.
\newblock Sun: A bayesian framework for saliency using natural statistics.
\newblock {\em Journal of Vision}, 8(7):32.1, 2008.

\end{thebibliography}
}

\end{document}